\documentclass{article}

\usepackage{microtype}
\usepackage{graphicx}
\usepackage{subfigure}
\usepackage{booktabs} % for professional tables

\usepackage{hyperref}

\usepackage[accepted]{icml2023}

\usepackage{amsmath}
\usepackage{amssymb}
\usepackage{mathtools}
\usepackage{amsthm}

\usepackage[font=small,skip=10pt]{caption}

\usepackage[capitalize,noabbrev]{cleveref}

\theoremstyle{plain}

\theoremstyle{definition}

\theoremstyle{remark}

\usepackage[textsize=tiny]{todonotes}

\icmltitlerunning{Submission and Formatting Instructions for ICML 2023}

\begin{document}

\twocolumn[
\icmltitle{Accelerating LLM Inference with Staged Speculative Decoding}

% It is OKAY to include author information, even for blind
% submissions: the style file will automatically remove it for you
% unless you've provided the [accepted] option to the icml2023
% package.

% List of affiliations: The first argument should be a (short)
% identifier you will use later to specify author affiliations
% Academic affiliations should list Department, University, City, Region, Country
% Industry affiliations should list Company, City, Region, Country

% You can specify symbols, otherwise they are numbered in order.
% Ideally, you should not use this facility. Affiliations will be numbered
% in order of appearance and this is the preferred way.
\icmlsetsymbol{equal}{*}

\begin{icmlauthorlist}
\icmlauthor{Benjamin Spector}{stanford}
\icmlauthor{Chris Re}{stanford}
%\icmlauthor{}{sch}
%\icmlauthor{}{sch}
\end{icmlauthorlist}

\icmlaffiliation{stanford}{Department of Computer Science, Stanford University, California, United States}

\icmlcorrespondingauthor{Benjamin Spector}{bfs@stanford.edu}

% You may provide any keywords that you
% find helpful for describing your paper; these are used to populate
% the "keywords" metadata in the PDF but will not be shown in the document
\icmlkeywords{Machine Learning, ICML}

\vskip 0.3in
]

% this must go after the closing bracket ] following \twocolumn[ ...

% This command actually creates the footnote in the first column
% listing the affiliations and the copyright notice.
% The command takes one argument, which is text to display at the start of the footnote.
% The \icmlEqualContribution command is standard text for equal contribution.
% Remove it (just {}) if you do not need this facility.

\printAffiliationsAndNotice{}  % leave blank if no need to mention equal contribution
% \printAffiliationsAndNotice{\icmlEqualContribution} % otherwise use the standard text.

\begin{abstract}
Recent advances with large language models (LLM) illustrate their diverse capabilities. We propose a novel algorithm, staged speculative decoding, to accelerate LLM inference in small-batch, on-device scenarios. We address the low arithmetic intensity of small-batch inference by improving upon previous work in speculative decoding. First, we restructure the speculative batch as a tree, which reduces generation costs and increases the expected tokens per batch. Second, we add a second stage of speculative decoding. Taken together, we reduce single-batch decoding latency by 3.16x with a 762M parameter GPT-2-L model while perfectly preserving output quality.
\end{abstract}

\section{Introduction}
\label{submission}

Large Language Models (LLMs) have witnessed tremendous growth over the last few years, demonstrating capabilities that range from high-quality text generation to complex reasoning, decision-making, and problem-solving tasks \cite{gpt3, gpt4, palm}. These strides, enabled by advances in deep learning architectures \cite{attentionisallyouneed}, training methodologies \cite{adam}, and vast amounts of data \cite{dataeffectiveness, thepile, thestack}, have paved the way for applications in fields as varied as natural language processing \cite{gpt3}, machine translation \cite{t5paper}, code synthesis \cite{codex}, and beyond \cite{gpt4}.

However, this exciting progress comes with its own set of system-level challenges. As LLMs have become more powerful, their computational demands have increased in tandem, often requiring substantial cloud resources for inference \cite{sheng2023high}. This requirement is prohibitive for many potential applications, especially those requiring low-latency responses \cite{wang2023tabi} or those where data privacy is paramount \cite{carlini2021extracting}.

Our paper addresses these challenges by accelerating local (small-batch) inference for LLMs, which suffers from poor compute utilization due to its low arithmetic intensity. We view this problem as crucial for three reasons: latency, personalization, and privacy. First, optimizing local inference latency improves real-time interactivity and responsiveness. Accelerating local inference also opens the door for more personalized LLM experiences as it allows models to be locally tailored to individual users. Finally, local inference enhances data privacy, as it removes the need for data to leave the user's device.

More philosophically, we believe that methods to efficiently run LLMs locally promote AI democratization by empowering individuals with limited computational resources. 

In this work, we build on the speculative decoding techniques introduced by \cite{matias, deepmindspec}, which use a fast but inaccurate draft model to anticipate the oracle model and batch queries to it, which improves sequential decoding performance while perfectly retaining the model distribution. These techniques scale well at first but their performance gains quickly saturate, because the probability of a draft model correctly guessing many sequential tokens is exponentially small. We improve speculative methods in two key ways:
\begin{enumerate}
    \item We restructure the speculative batch as a tree of possible token sequences, so as to more quickly create larger and higher quality speculative batches.
    \item We speculatively decode the draft model, too, to further improve performance.
\end{enumerate}

We find these techniques significantly improve the performance of speculative decoding in both deterministic and sampling-based decoding.

\section{Background}

In this section, we provide a brief overview of autoregressive LLM inference, key principles of GPU performance optimization, and prior work in optimizing LLM inference.

\subsection{Autoregressive LLM Inference}

Autoregressive generation from decoder-only LLMs is generally split into two phases. First, the prompt is run through the model to generate the KV cache and the first output logits. This is usually fast, as the entire prompt can be handled in parallel.

The second phase is decoding. A token is selected from the outputted logits and fed back into the model, which produces logits for the following token. This is repeated until the desired number of tokens is produced. Because decoding must be done sequentially, with the entire model’s weights streamed through the compute units each time in order to generate a single token, the arithmetic intensity (that is, FLOP of compute / byte of memory bandwidth) of this second phase is extremely low when run in small batches. As such, decoding is usually the most expensive part of autoregressive generation. \citep{matias}

\subsection{GPU optimization}

Modern LLM inference is most often conducted on GPUs due to the highly parallel nature of the workload, which consists principally of large matrix multiplications.

\begin{figure}
    \centering
    \includegraphics[width=0.5\textwidth]{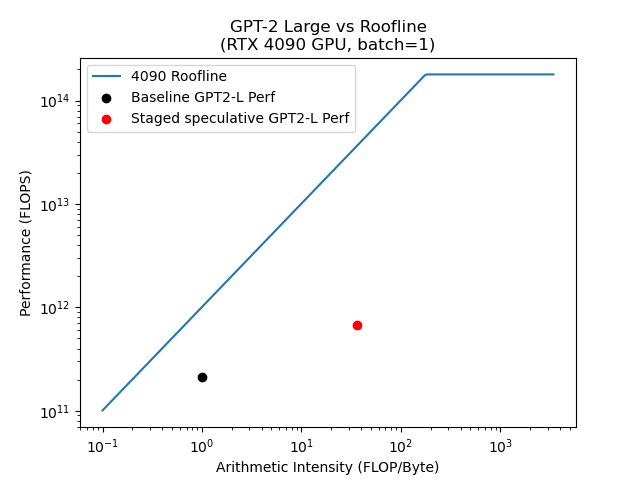}
    \caption{A roofline plot for single-query GPT-2-L inference on an RTX 4090. At small batch sizes, inference is completely memory bandwidth bound. Thus this plot shows that the only way to significantly increase performance is to increase the arithmetic intensity of inference.}
    \label{fig:roofline}
\end{figure}

GPUs consist of thousands of extremely small efficient cores supported by a multi-level memory hierarchy. The key challenge of optimizing small-batch LLM inference for GPUs is to deal with the extremely low arithmetic intensity. Operating in 16-bit precision with a batch size of 1, decoding has an arithmetic intensity of 1. For example, for a reference PyTorch \cite{paszke2019pytorch} implementation of GPT-2 Large (762M parameters), inference requires approximately 1.4 GFLOP, and yet a quiesced NVIDIA RTX 4090 achieves only 150 tokens/second, for a compute utilization of a mere 0.13\% \cite{nvidia2022nvidia}. This abysmal performance is substantially due to the GPU roofline \cite{ofenbeck2014applying}, which is governed by memory bandwidth at low arithmetic intensities (visualized in Figure \ref{fig:roofline}).

\subsection{Speculative Decoding}

There are many techniques under investigation today to accelerate inference, such as quantization \cite{dettmers2022llm,frantar2022gptq}, flash attention \cite{dao2022flashattention}, and speculative decoding \cite{matias, deepmindspec}. In this section, we'll briefly examine speculative decoding as described in \cite{matias, deepmindspec}, as it is the primary subject of this work.

The basic idea of speculative decoding is to use a smaller, faster draft model to decode several tokens in advance, and then feeds them into the oracle model as a single batch. If the draft model was right about its predictions -- the larger model agrees -- one can decode several tokens with a single batch, which saves considerable memory bandwidth, and thus time, per token. However, if the larger model rejects the tokens predicted by the draft model, then the rest of the batch is discarded and the algorithm naturally reverts to standard token-by-token decoding. Speculative decoding may also be accompanied by a rejection sampling scheme to sample from the original distribution. Note this is only useful in small-batch settings where bandwidth is the bottleneck. Speculative decoding trades compute for bandwidth.

There are two key reasons why speculative decoding is an attractive performance engineering target. First, it does not degrade model quality at all. Second, the gains it provides are generally orthogonal to other methods, because its performance comes from converting sequential execution to parallel execution. \citep{matias}

\section{Methods}

We make two improvements to speculative decoding: tree-structured batches, and additional stages. We term the combination of these methods ``staged speculative decoding''.

\subsection{Tree-structured batches}

Current speculative methods predict a single sequence for the batch. However, this doesn’t scale well to large batch sizes or low draft model alignments. Intuitively, the probability that two models agree for long consecutive sequences of tokens is exponentially low, which means that speculative decoding has rapidly diminishing returns as one scales its arithmetic intensity.

Our approach is to dynamically build a tree of the possible sequences, which provides three benefits: more expected true tokens per batch, an increased number of leaf nodes, and better parallelism for the small draft model.

\begin{table}[]
    \centering
    \resizebox{\columnwidth}{!}{%
    \begin{tabular}{c|c|c|c}
        Sampling & Baseline & Speculative & Staged spec.\\
        method & rel. bandwidth & rel. bandwidth & rel. bandwidth \\
        \hline
        \hline
        Deterministic & $1.00$ & $0.31$ & $\mathbf{0.23}$ \\
        \hline
        Topk & $1.00$ & $0.48$ & $\mathbf{0.35}$
    \end{tabular}
    }
    \caption{Memory bandwidth consumption (relative to baseline) of speculative and staged speculative decoding methods.}
    \label{tab:mem}
\end{table}

First, by reallocating computation from the end of very long sequences to the beginning, and considering the second or third most likely tokens to be produced by the model, one increases the expected number of tokens per batch compared to the naive approach.

Second, the cost of running the draft model to produce the batch is non-negligible in standard speculative decoding. However, in a tree of predictions which constitute the batch to the oracle model, the draft model is only run at internal nodes of the tree. So, a wider tree increases the number of leaf nodes, which means that one gets more of the batch for free.

A third benefit of the wider tree is that one can parallelize execution for the small model across the tree, which also decreases its cost. In the limit, one only needs to run the draft on a number of batches equal to the depth of the tree. This is important because draft models are usually smaller transformer-based models and are thus memory-bound in small-batch inference, too.

Implementing a tree-structured batch requires some care. The simplest approach is to partition self-attention while decoding into cross-attention with the KV cache and self-attention within the batch. The tree-structured batch can then be constructed by controlling both the positional embeddings and causally masking the batch self-attention matrix according to the tree. Finally, the new KV cache for the whole batch must be stored separately, and then the appropriate slices appended to the main KV cache after tokens are sampled.

\subsection{Staged Speculation}

Current speculative methods use a single smaller model as the draft, usually a smaller LLM \cite{deepmindspec}. In this setting, the size of the draft model is an important hyperparameter: a larger draft model will have better alignment with the oracle, but will cost more, whereas a smaller model will produce lower quality speculative batches, but at a lower cost. In practice, draft models that are about 15-20x smaller than the oracle seem optimal. 

However, under naive speculative decoding, assembling large batches inverts the cost structure, with more time spent on the draft model than the oracle. So, one should accelerate the draft model in generating sequences of tokens, and speculative decoding is a natural solution for this, too. We correspondingly add speculative decoding to the draft model in our approach. Thus the overall method of ``staged speculative decoding'', consists of oracle, draft, and draft$^2$ models with tree-structured batches.

\section{Results}

For our experiments, we use three models: a GPT-2-Large (762M) parameter oracle model \cite{gpt2} fine-tuned on the Python subsection of the Stack \cite{thestack}, a small (40M) parameter GPT-2 draft model trained on the same, and a Katz backoff trigram model \cite{katzmodel} as the draft$^2$ model. The Katz backoff model was generated by running the draft model for two hours at a sampling temperature of 1.5 to generate 120M tokens. All evaluations were conducted on a quiesced RTX 4090 GPU \cite{nvidia2022nvidia}, which is top-end consumer hardware.

We evaluate against two alternative inference methods. First, our standard baseline is simple token-by-token decoding with the oracle. Second, we also evaluate against speculative decoding as proposed by \cite{matias}, so as to isolate the effects of our improvements.

\begin{table}[]
    \centering
    \resizebox{\columnwidth}{!}{%
    \begin{tabular}{c|c|c|c}
        Sampling & Baseline & Speculative & Staged spec.\\
        method & tokens/sec & tokens/sec & tokens/sec \\
        \hline
        \hline
        Deterministic & $150$ & $350$ & $\mathbf{475}$ \\
        \hline
        Topk & $150$ & $219$ & $\mathbf{298}$
    \end{tabular}
    }
    \caption{Relative performance (in tokens/second decoded) with baseline (non-speculative), standard speculative, and staged speculative decoding methods.}
    \label{tab:performance}
\end{table}

To evaluate, we ran the 164 prompts from HumanEval \cite{codex}, using non-speculative, speculative, and our staged speculative methods, and with both deterministic and topk sampling \cite{gpt2}. Details of batch sizes and internal heuristics can be found in our \href{https://anonymous.4open.science/r/stagedspeculation-988F/}{code}.

We first measured the memory bandwidth requirements of each method, to validate that our approach saves appreciable bandwidth. We detail the results in table \ref{tab:mem}, which illustrate that staged speculative decoding uses substantially less memory bandwidth than either alternative method.

\begin{figure}
    \centering
    \includegraphics[width=\columnwidth]{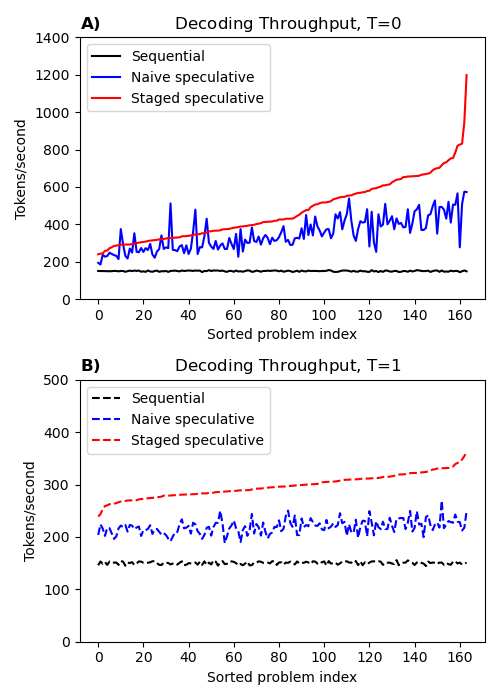}
    \caption{Relative performance distribution over different problems in the HumanEval dataset. (A) shows greedy decoding, whereas (B) shows Topk decoding. Problem indices are sorted by staged speculative performance for clarity.}
    \label{fig:relativeperf}
\end{figure}

Second, we measure sequential decoding throughput for each approach. The results are summarized in table \ref{tab:performance} and detailed in Figure \ref{fig:relativeperf}.

With deterministic sampling, our implementation provides an average performance boost of 3.16x over our reference implementation, and 1.36x over standard speculative sampling. Furthermore, we evaluate on relatively small models, whereas prior work uses much larger models on which one would expect greater benefits. Profiling data shows our implementation has 35\% overhead from the Python infrastructure, which could be reduced by a more efficient implementation or amortized over larger models.

With topk ($k=50,T=1$) sampling, although both speculative methods are significantly degraded due to stochastic rejection of tokens provided in the batch, staged speculation nonetheless retains its lead, providing an average performance boost of 1.98x over baseline and 1.36x again over standard speculative sampling.

\begin{figure}
    \centering
    \includegraphics[width=\columnwidth]{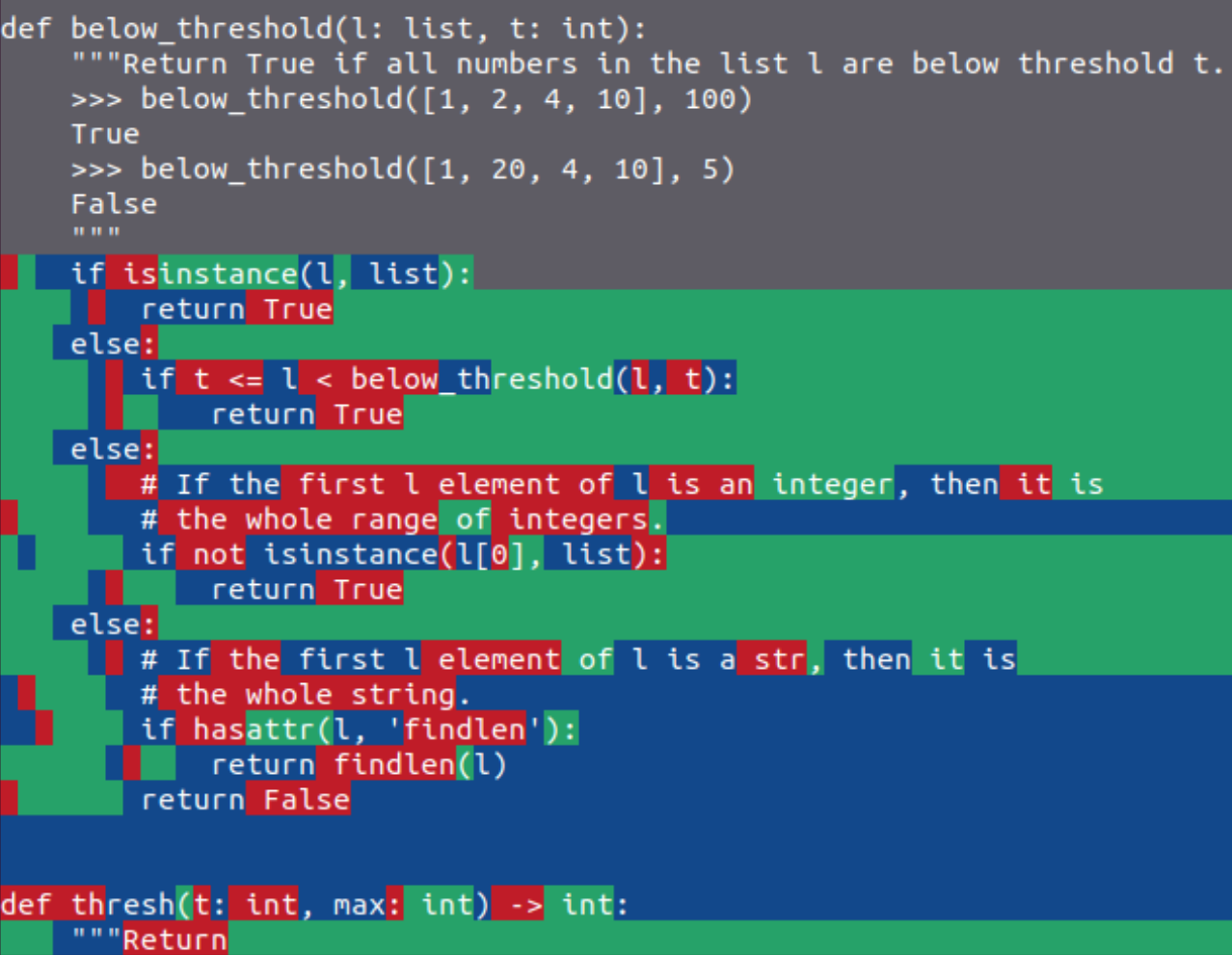}
    \caption{A visualization of the origin of tokens in an example T=1 HumanEval completion. Green background originates with the N-gram draft$^2$ model, blue the draft model, and red the oracle model. (Of course, all tokens are eventually checked by the oracle model.) Obvious tokens -- like whitespace -- are preferentially accelerated relative to difficult ones.}
    \label{fig:origins}
\end{figure}

In Figure \ref{fig:origins}, we show the origin of different tokens in the completed model. (The performance gain on the shown prompt is approximately 2.5x over baseline.) The model is usually able to decode the easiest, most obvious tokens, like whitespace, in batch through both transformer models, as they originate with the N-gram models. Somewhat more difficult tokens are generated by the small model, while the most critical tokens (like the token following the “if” token) come from the oracle model. Note that due to the finite batch size, the above is only a trend and should not be expected to apply universally to every token. Some tokens which could have been accurately predicted by a smaller model will still end up originating from larger models.

We also wish to acknowledge the extreme range of the performance benefits as a downside of the work. While performance benefits run as high as 10x on realistic prompts, they can also be limited to only 2x. To a large degree, this depends on the denseness or sparseness of difficult content. For example, highly indented Python code will make better use of the N-gram models than unindented code, and thus reap greater performance benefits.

We speculate that these models represent an approximately fixed cost per entropy of the data. Extremely low entropy generation, like pure whitespace, will be generated very quickly by staged speculative decoding, with performance approaching that of large-batch inference, whereas dense generations with high entropy will need to rely on small-batch decoding at all stages. So, a corollary implication of this work is that most of the text generated by LLMs has entropy lower than the capabilities of their authoring models, and that the increased accuracy of big models is isolated to a relatively small number of key tokens.

We see several paths for future work:
\begin{enumerate}
    \item We suspect it may be possible to speculatively sample with $T>0$ even faster by generating the multinomial CDFs first, and then using this sequence to help choose the tokens to assemble into the full batch. For example, if the multinomial CDF sampled is $0.99$, it may be best to only include in the batch the draft model's fifth through tenth most likely tokens.
    \item Running with larger models would likely yield even greater performance boosts while still fitting on-device. With 8-bit quantization, it should be possible to fit 20B models on consumer GPUs in small-batch, allowing for an entire additional stage of speculation.\\($20B \rightarrow 1B \rightarrow 50M \rightarrow \text{N-gram}$).
    \item Investigating better lowest-level draft models could also improve performance -- models which perform better than N-gram models but still run in $<10\mu s$.
\end{enumerate}

\section{Conclusions}

In this work, we described and implemented several improvements over previous work in speculative decoding. First, we restructured the batch provided to the oracle model as a tree, in order to decrease the cost of generation and increase the expected number of tokens per batch. Second, we added a second stage of speculation to accelerate the decoding of the draft model. Altogether, we achieved an average speedup of 3.16x over standard single-batch inference.

% Acknowledgements should only appear in the accepted version.
\section*{Acknowledgements}

[Left blank for blind review.]

% We thank Laurel Orr for providing the GPT-2 models fine-tuned on Python. Additional thanks to Tri Dao, Dan Fu, and Beidi Chen for helpful conversations, and Tri and Dan in particular for their Flash Attention code, which was an important reference for this work.

% In the unusual situation where you want a paper to appear in the
% references without citing it in the main text, use \nocite
\nocite{langley00}

\bibliography{example_paper}
\bibliographystyle{icml2023}

%%%%%%%%%%%%%%%%%%%%%%%%%%%%%%%%%%%%%%%%%%%%%%%%%%%%%%%%%%%%%%%%%%%%%%%%%%%%%%%
%%%%%%%%%%%%%%%%%%%%%%%%%%%%%%%%%%%%%%%%%%%%%%%%%%%%%%%%%%%%%%%%%%%%%%%%%%%%%%%
% APPENDIX
%%%%%%%%%%%%%%%%%%%%%%%%%%%%%%%%%%%%%%%%%%%%%%%%%%%%%%%%%%%%%%%%%%%%%%%%%%%%%%%
%%%%%%%%%%%%%%%%%%%%%%%%%%%%%%%%%%%%%%%%%%%%%%%%%%%%%%%%%%%%%%%%%%%%%%%%%%%%%%%
% \newpage
% \appendix
% \onecolumn
% \section{You \emph{can} have an appendix here.}

% You can have as much text here as you want. The main body must be at most $8$ pages long.
% For the final version, one more page can be added.
% If you want, you can use an appendix like this one, even using the one-column format.
%%%%%%%%%%%%%%%%%%%%%%%%%%%%%%%%%%%%%%%%%%%%%%%%%%%%%%%%%%%%%%%%%%%%%%%%%%%%%%%
%%%%%%%%%%%%%%%%%%%%%%%%%%%%%%%%%%%%%%%%%%%%%%%%%%%%%%%%%%%%%%%%%%%%%%%%%%%%%%%

\end{document}